\documentclass[sigconf]{acmart}
\AtBeginDocument{%
  }

\usepackage{graphicx}
\usepackage{url}

\copyrightyear{2026}
\acmYear{2026}
\setcopyright{cc}
\setcctype{by}
\acmConference[CIKM '26]{Proceedings of the 35th ACM International Conference on Information and Knowledge Management}{November 07--11, 2026}{Rome, Italy}
\acmBooktitle{Proceedings of the 35th ACM International Conference on Information and Knowledge Management (CIKM '26), November 07--11, 2026, Rome, Italy}
\acmDOI{10.1145/3799682.3840182}
\acmISBN{979-8-4007-2539-5/2026/11}

\begin{document}

\title{RedditPersona: A Modular Framework for Community-Conditioned LLM Adaptation from Reddit}

\author{Amirhossein Ghaffari}
\orcid{0009-0006-9264-8681}
\affiliation{%
  \institution{Future Computing Group\\ University of Oulu}
  \city{Oulu}
  \country{Finland}
}
\email{amirhossein.ghaffari@oulu.fi}

\author{Ali Goodarzi}
\orcid{0009-0005-1578-3700}
\affiliation{%
  \institution{Centre for Applied Computing\\ University of Oulu}
  \city{Oulu}
  \country{Finland}
}
\email{ali.goodarzi@oulu.fi}

\author{Huong Nguyen}
\orcid{0000-0001-9067-3396}
\affiliation{%
  \institution{Future Computing Group\\ University of Oulu}
  \city{Oulu}
  \country{Finland}
}
\email{huong.nguyen@oulu.fi}

\author{Simo Hosio}
\orcid{0000-0002-9609-0965}
\affiliation{%
  \institution{Centre for Applied Computing\\ University of Oulu}
  \city{Oulu}
  \country{Finland}
}
\email{simo.hosio@oulu.fi}

\author{Lauri Lovén}
\orcid{0000-0001-9475-4839}
\affiliation{%
  \institution{Future Computing Group\\ University of Oulu}
  \city{Oulu}
  \country{Finland}
}
\email{lauri.loven@oulu.fi}

\author{Ekaterina Gilman}
\orcid{0000-0001-9816-2240}
\affiliation{%
  \institution{Future Computing Group\\ University of Oulu}
  \city{Oulu}
  \country{Finland}
}
\email{ekaterina.gilman@oulu.fi}

\renewcommand{\shortauthors}{Ghaffari et al.}

\begin{abstract}
Community-conditioned language model adaptation needs choices about data collection, community definition, and evaluation that are currently made independently in each study, making it hard to compare assumptions or reuse artifacts. We present \textsc{RedditPersona}, a modular framework that standardizes these choices: it collects Reddit posts and comments, profiles active users, partitions them under five grouping strategies (subreddit-based, graph-structural, semantic, hybrid, and interaction-based), trains a parameter-efficient adapter per strategy via QLoRA, and evaluates them under a shared metric suite spanning fluency, fidelity, distributional alignment, and community identifiability. Applied to 112 subreddits in the urban well-being domain (301{,}429 user profiles, 16M+ comments), we find that adapters' behavioral identifiability tracks each strategy's agreement with the subreddit baseline, and that a consistent trade-off between identifiability and distributional similarity to real text holds across all five strategies. The code and configuration files are available at: \url{https://github.com/Ahghaffari/redditpersona}.
\end{abstract}




\maketitle

\section{Introduction}
\label{sec:intro}
Online communities provide one of the richest sources of behavioral traces for studying how people express preferences, identities, emotions, and social roles. At the same time, large language models (LLMs) are increasingly used not only to analyze such traces but also to simulate users, groups, and online discussions. Recent studies show that LLMs can generate realistic social media conversations~\cite{bouleimen2025collective}, imitate polarized political comments after fine-tuning~\cite{vendetti2025passing}, model group identity through community-specific data~\cite{torres2026phenomenologically}, and support large-scale social or policy simulations~\cite{dong2025simulating,huang2026policysim,malvicini2026natural}. These developments create a growing need for resources that enable reproducible, comparable, and reusable social media-based personalization and community-level adaptation.

However, current work often treats data construction, community definition, model adaptation, and evaluation as tightly coupled choices made for one study. A dataset may be collected for one domain, a grouping method may be fixed by design, or a fine-tuning script may assume a single model family or prompt format. This makes it difficult to answer basic methodological questions: whether a subreddit boundary, an interaction graph, or a semantic cluster is the most appropriate unit; whether the same users can be reorganized for different research questions; and whether community-conditioned models are genuinely different from a global baseline. The problem is especially important for community-level alignment, which has been proposed as a middle ground between one-size-fits-all alignment and costly individual-level personalization~\cite{lin2026communitybench}. Without a standard pipeline, researchers must repeatedly rebuild scrapers, profilers, clustering code, training-data converters, and evaluation scripts before they can compare modeling assumptions.

We introduce \textsc{RedditPersona}, a modular resource for turning raw Reddit posts and comments into reusable community-conditioned LLM fine-tuning and evaluation artifacts. The framework is designed around a simple principle: the same raw social activity should support multiple downstream definitions of ``community'' and should remain reusable until the fine-tuning-ready stage. Starting from a user-specified set of subreddits, \textsc{RedditPersona} collects posts and comments, profiles active users, constructs activity and interaction artifacts, applies multiple community grouping strategies, analyzes the resulting partitions with shared metrics, emits instruction-tuning data in a portable chat format, and trains parameter-efficient adapters per grouping strategy.

As a case study, we focus on subreddits related to urban well-being. Well-being is a multidimensional construct shaped by social, environmental, and urban factors~\cite{ghaffari2025understanding}, making it a useful domain for testing whether different grouping assumptions reveal different community contexts. This paper makes three contributions:

\begin{itemize}
    \item First, we release an end-to-end, configurable framework that standardizes Reddit data collection, user profiling, community construction, training data generation, adapter fine-tuning, and evaluation.
    \item Second, we provide a unified interface for comparing multiple community grouping strategies with common post-processing and metrics.
    \item Third, we provide reusable scripts, including configuration, anonymization, fine-tuning, and an evaluation benchmark, so that future work can compare community-conditioned LLMs without rebuilding the pipeline.
\end{itemize}

\section{Related Work}
\label{sec:related}
\paragraph{LLM-based social simulation and user modeling.}
LLMs are increasingly used to simulate individuals, groups, and online communities. Early work in this direction has examined whether LLM-generated discussions can resemble real social media conversations~\cite{bouleimen2025collective}, whether LLM agents can support social network simulation~\cite{dong2025simulating,jiang2025social}, and whether simulation sandboxes can be used to study policy interventions before deployment~\cite{huang2026policysim}. Other studies focus on specific social phenomena, such as affective polarization~\cite{malvicini2026natural}, emotion diffusion~\cite{qiang2025emotion}, political persuasion~\cite{bai2025llm}, and the simulation of judgment~\cite{loru2025simulation}. These works demonstrate the promise of LLM-based simulation, but they also show that validity depends strongly on how users, communities, and interaction histories are represented.

\paragraph{Personalization, personas, and community-level alignment.}
A parallel line of work studies how LLMs can represent individual users or population groups. Fine-tuning has been shown to improve behavioral prediction in social science experiments~\cite{kolluri2025finetuning}, while HumanLM argues that user simulation should align with latent states such as beliefs, emotions, values, and communication style rather than only surface text imitation~\cite{wu2026humanlm}. Other work evaluates personalized role-playing on Reddit-like social media data~\cite{li2026imitation} and conditioned comment prediction for social media users~\cite{schwager2026towards}. Persona-based population simulation and population-aligned persona generation further emphasize that collections of personas must reflect target populations, not only plausible individual profiles~\cite{hu2025population,li2026persona}. At the group level, CommunityBench frames community-level alignment as a scalable alternative between global and individual alignment~\cite{lin2026communitybench}. \textsc{RedditPersona} complements these efforts by providing the data and training pipeline needed to construct and compare community-conditioned adapters under multiple definitions of community.

\paragraph{Fine-tuning LLMs for social identity and personality.}
Several studies show that LLM behavior can be steered through fine-tuning or parameter-efficient adaptation, including mimicking polarized political discourse~\cite{vendetti2025passing}, simulating online group identity~\cite{torres2026phenomenologically}, and a range of personality-related tasks~\cite{Shen2025LessBB,Hu2024LLMVS,Zhu2025EvaluatingLA,Zhan2025TestTimeMatchingDP,banerjee2025fine}. These studies motivate the need for a reusable framework.

\paragraph{Datasets and resources for social media agents.}
Recent resources have begun to standardize parts of the social-media simulation workflow. BluePrint and SIMPACT provide privacy-preserving social media user datasets for persona evaluation and training~\cite{BluePrint2025}. Other work constructs Reddit-based datasets for agent-based modeling with users, topics, and interaction networks~\cite{sittar2026}. These resources are valuable, but they typically fix a particular platform, task formulation, or community abstraction. \textsc{RedditPersona} is instead designed as a framework for producing reusable artifacts from new subreddit collections: raw JSONL files, user profiles, activity matrices, interaction graphs, alternative community assignments, fine-tuning-ready conversations, adapters, and evaluation tables. In this way, it acts as an experimental pipeline for testing how data grouping choices affect community-conditioned LLM behavior.

\section{Framework}
\label{sec:framework}

\begin{figure*}
    \centering
    \includegraphics[width=0.93\textwidth]{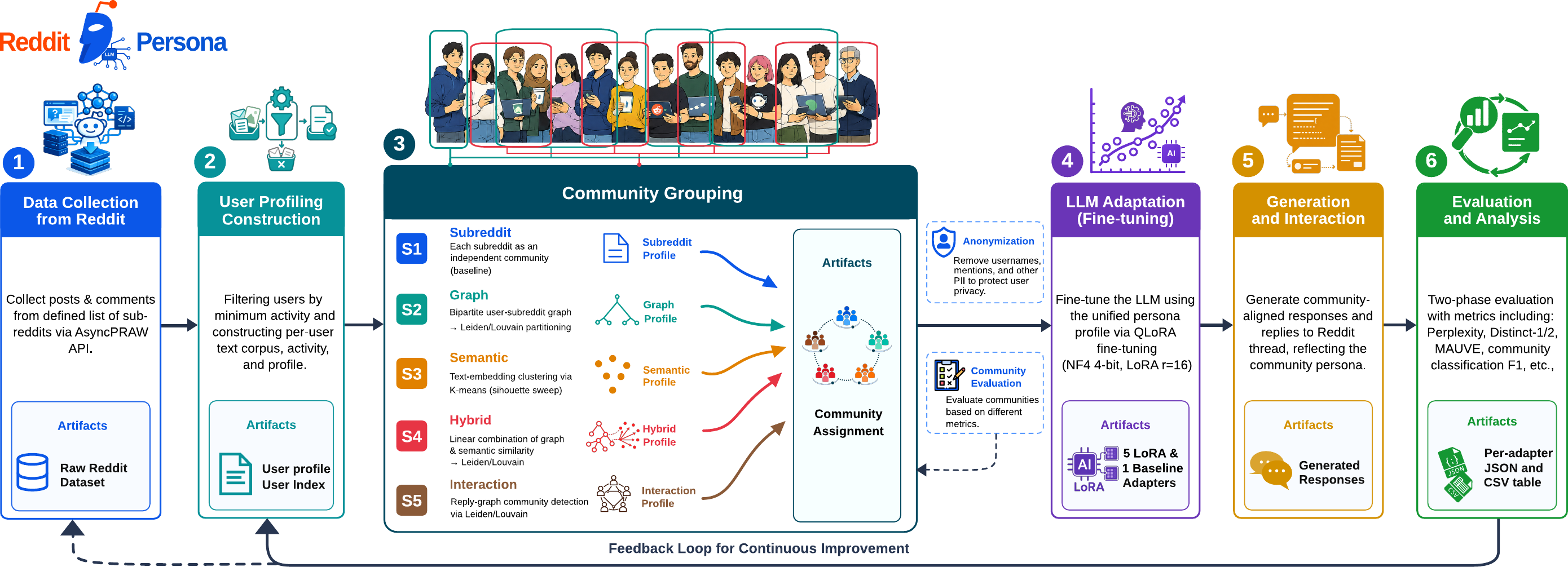}
    \caption{\textsc{RedditPersona} pipeline}
    \label{fig:workflow}
\end{figure*}

\textsc{RedditPersona} is a modular framework that turns raw Reddit posts and comments into community-conditioned language model adapters, along with metrics for comparing them. Figure~\ref{fig:workflow} summarises the six phases and the artifacts each one produces. Every phase is exposed as an independent sub-command of a single CLI and is configurable through a layered configuration mechanism, so a user can re-run a single phase, swap a base LLM, change the target number of communities, or enable anonymization. A single execution uses one selected grouping strategy to partition the corpus and fine-tune adapters on that partition; to benchmark grouping effects, we keep the input data fixed and rerun the pipeline across all strategies, producing one adapter per strategy. The remainder of this section walks through the phases at a high level; parameter details and reproducibility scripts are available in the repository.

\paragraph{Phase 1 — Data collection from Reddit.}
The user supplies a list of subreddits, optionally organized into thematic categories for a specific study. A verification step filters out private, Not Safe For Work (NSFW), or low-subscriber communities before collection begins. An \textsc{AsyncPRAW}-based scraper can then mix \textit{hot}, \textit{top}, and \textit{new} sort orders, apply exponential back-off on the rate limits, and checkpoints every $N$ submissions to allow resumption. Posts and comments are streamed to per-subreddit JSONL files, while two compact relational artifacts are materialized on the fly: a sparse user~$\times$~subreddit \emph{activity matrix} and a directed user~$\to$~user \emph{interaction graph} of reply edges weighted by frequency.

\paragraph{Phase 2 — User profiling construction.}
A single-pass streaming aggregator builds, for every user passing a configurable minimum activity filter, three artifacts: a structured profile (subreddit histogram, first/last activity, totals), an activity matrix slice, and a concatenated text corpus written through a bounded LRU pool of file handles so that large user populations can be processed without exhausting RAM or file descriptors.

\paragraph{Phase 3 — Community grouping.}
Five community detection strategies share a uniform interface: (\textbf{S1})~the \emph{subreddit baseline}, where each subreddit is its own multi-membership community; (\textbf{S2})~a \emph{graph} strategy that projects the bipartite user-subreddit graph into a sparse user-user similarity graph and partitions it with Leiden~\cite{traag2019louvain} (Louvain~\cite{blondel2008fast} fallback); (\textbf{S3})~a \emph{semantic} strategy that embeds each user's text corpus and clusters via $K$-means~\cite{pedregosa2011scikit} with a silhouette~\cite{dudek2019silhouette} over configurable $K$; (\textbf{S4})~a \emph{hybrid} strategy that linearly blends graph and cosine semantic similarity on existing edges and re-applies Leiden (Louvain fallback); and (\textbf{S5})~an \emph{interaction} strategy that runs Leiden~\cite{traag2019louvain} on the reply graph with a Louvain fallback. Since Louvain and Leiden produce long-tailed partitions, a shared post-processing step consolidates each strategy's output into its top-$K$ largest communities plus an \textit{other} class, yielding comparable cardinalities across strategies.

An anonymization stage rewrites usernames in place using a salted HMAC-SHA256 hash, strips URLs, and removes personal-name PII detected by spaCy \cite{spacy}. 

A dedicated community analyzer computes, for every strategy, a set of metrics: community-size distribution and Gini coefficient, intra-community coherence and inter-community separation on the pretrained \texttt{Google EmbeddingGemma-300M} centroid cosine, TF-IDF vocabulary distinctiveness \cite{qaiser2018text}, and Normalized Mutual Information (NMI) and Adjusted Rand Index (ARI) against the subreddit baseline (S1) as the reference grouping.

\paragraph{Phase 4 — LLM adaptation (fine-tuning).}
For each grouping strategy, any HuggingFace instruction-tuned LLM can be fine-tuned with QLoRA~\cite{dettmers2023qlora}: 4-bit NF4 quantization with double-quant, fp16 compute, and LoRA~\cite{hu2022lora} adapters on attention projections, driven by TRL's \texttt{SFTTrainer}~\cite{vonwerra2020trl}. Training data is emitted in the OpenAI \texttt{messages} format, with the community identity encoded in the system prompt; the tokenizer's own chat template is automatically applied at training time, making the same training script portable across model families (Qwen, Llama, Mistral, Gemma, etc.). One \emph{community-pooled} adapter is trained per strategy, plus a strategy-agnostic \emph{baseline\_all} adapter, all written to a deterministic on-disk layout for downstream evaluation.

\paragraph{Phase 5 — Generation and interaction.}
The pipeline loads each trained adapter and the base model for zero-shot baselines, and generates replies for held-out test conversations using the same tokenizer and chat template as during training.

\paragraph{Phase 6 — Evaluation and analysis.}
Phase~A computes per-token perplexity (PPL) using the same chat template as training. Phase~B scores the generated text against references and against the corresponding community corpus, combining lexical metrics (Distinct-$n$ (Dist-$n$)~\cite{li2016diversity}, TF-IDF vocabulary Jaccard (Vocab-Jacc)~\cite{qaiser2018text}), semantic metrics (BERTScore-F1~\cite{zhang2019bertscore}, MAUVE~\cite{pillutla2021mauve}), distributional metrics (LDA Topic KL Divergence (Topic-KL)~\cite{jelodar2019latent}, VADER sentiment Jensen-Shannon Divergence (Sent-JSD)~\cite{hutto2014vader}), and a discriminative Community-F1 probe (Comm-F1) that trains a TF-IDF logistic-regression classifier on generated text to recover the originating community label.

\section{Experimental Evaluation}
\label{sec:eval}

\subsection{Experimental Setup and Benchmark}
\begin{table}[t]

\centering
\caption{Reddit corpus statistics.}

\label{tab:dataset}
\small
\resizebox{0.8\columnwidth}{!}{
\begin{tabular}{lr}
\toprule
\textbf{Statistic} & \textbf{Count} \\
\midrule
Subreddits collected       & 112           \\
Posts                             & 152{,}761         \\
Comments                                         & 16{,}061{,}338    \\
Unique users (with profiles)                     & 301{,}429         \\
Author-reply edges in interaction graph         & 3{,}660{,}079     \\
User-user co-participation edges (S2/S4 graph)  & 186{,}305{,}455   \\
Time span of collected content                   & 2022-01 -- 2026-04 \\
\bottomrule
\end{tabular}
}
\end{table}

Table~\ref{tab:dataset} summarises the corpus. We collected 112 subreddits across 14 thematic categories of urban well-being plus one cross-cutting category~\cite{ghaffari2025understanding} using \textsc{AsyncPRAW} with \textit{hot}, \textit{top} (last year only), and \textit{new} sort orders. Users were retained for community construction only if they contributed at least ten comments across at least two distinct subreddits, yielding 301{,}429 profiles.

\paragraph{Community grouping.}
All strategies (Section~\ref{sec:framework}) were run on the full corpus. Partitions were consolidated to the top-$K$ largest communities, with smaller-community members merged into an ``other'' bucket excluded from metrics: $K{=}112$ for S1 (one community per subreddit, where each user is assigned to their highest-activity subreddit), $K{=}100$ for S2 and S4 (Leiden, resolution~2.0), $K{=}100$ for S5 (Leiden, resolution~1.0), and $K{=}80$ for S3 (silhouette-optimal $k \in \{80, 90, 100, 110, 120\}$). Text embeddings for S3 and S4 used \texttt{Google EmbeddingGemma-300M}. For training and evaluation, each strategy is restricted to the top-10 communities by size.

\paragraph{Training.}
For each grouping strategy, a single \texttt{IBM Granite 4.1-3B} adapter is fine-tuned with QLoRA~\cite{dettmers2023qlora} on the pooled data of all retained communities, with community identity encoded in the system prompt so the model learns to condition on it. Quantization uses 4-bit NF4 with double quantization and fp16 compute. LoRA~\cite{hu2022lora} adapters ($r{=}16$, $\alpha{=}32$, dropout~0.05) target all attention projections. Training uses TRL \texttt{SFTTrainer}~\cite{vonwerra2020trl} for 1~epoch, per-device batch size~2, gradient accumulation~8 (effective batch~16), learning rate $1{\times}10^{-5}$, cosine schedule, warmup ratio~0.10, and weight decay~0.01. Each sample is a reply thread in the OpenAI \texttt{messages} format: a system prompt declaring the assigned community and grouping strategy, a single user turn concatenating the post (title and body), up to three level comments, and the parent comment, and the community member's reply as the assistant turn, truncated to 512~tokens. Data is split 80/10/10. Two references are included: \textit{baseline\_all} pools all community data into a single adapter ($K{=}1$); \textit{zero\_shot} uses the unmodified base model.

\paragraph{Evaluation.}
Up to 200~held-out test prompts per adapter are used to generate replies (temperature~0.7, top-$p$~0.9, max 256~new tokens). We compute the nine metrics defined in Phase~6 (Section~\ref{sec:framework}): PPL, Dist-1/2, Vocab-Jacc, Topic-KL, Sent-JSD, BERTScore-F1, MAUVE, and Comm-F1. MAUVE is computed once per strategy by pooling generations and references across all evaluated communities ($2{,}000$ samples per side for adapter strategies, $1{,}000$ for zero-shot, $200$ for \textit{baseline\_all}), which keeps it within its recommended sample regime; Comm-F1 is defined only for community-conditioned adapters and is undefined for the \textit{baseline\_all} and \textit{zero\_shot} rows.

\subsection{Results and Discussion}
\begin{table}[t]
\centering
\caption{Community grouping metrics across strategies. Arrows lower ($\downarrow$) / higher ($\uparrow$) is better.}
\label{tab:grouping}
\resizebox{\columnwidth}{!}{%
\begin{tabular}{lrrrrrrr}
\toprule
Strategy & $K$ & Gini & Mean size & NMI vs.~S1 & ARI vs.~S1 & Intra-coh.~$\uparrow$ & Inter-sep.~$\downarrow$ \\
\midrule
S1 Subreddit (baseline)         & 112 & 0.593          & 2691.3 & —    & —    & 0.536          & 0.196 \\
S2 Graph (Leiden)               & 100 & 0.744          & 1284.7 & \textbf{0.586} & \textbf{0.251} & \textbf{0.567} & \textbf{0.137} \\
S3 Semantic ($k$-means)         &  80 & \textbf{0.211} & 3767.9 & 0.319 & 0.100 & 0.561          & 0.214 \\
S4 Hybrid (graph $\oplus$ sem.) & 100 & 0.716          & 1284.6 & 0.522 & 0.236 & 0.562          & 0.137 \\
S5 User-Interaction (Leiden)    & 100 & 0.929          & 3005.6 & 0.327 & 0.129 & 0.489          & 0.160 \\
\bottomrule
\end{tabular}
}
\end{table}

Table~\ref{tab:grouping} shows the community grouping comparison across strategies. Content-based partitions (S2, S4) yield the tightest clusters but inherit the long-tail size distribution, while S3 is the only strategy that produces near-uniform clusters at the cost of inter-cluster separation. S5 is most useful as a sociological control: its low coherence shows it captures social ties orthogonal to content.

\begin{table}[t]
\centering
\caption{Generation metrics across community-conditioned strategies. Arrows lower ($\downarrow$) / higher ($\uparrow$) is better.}
\label{tab:strategy_comparison}
\resizebox{\columnwidth}{!}{
\begin{tabular}{lcccccccccc}
\toprule
\textbf{Strategy} & \textbf{K} & \textbf{PPL $\downarrow$} & \textbf{Dist-1} & \textbf{Dist-2} & \textbf{Vocab-Jacc $\uparrow$} & \textbf{Topic-KL $\downarrow$} & \textbf{Sent-JSD $\downarrow$} & \textbf{BERTScore-F1 $\uparrow$} & \textbf{MAUVE $\uparrow$} & \textbf{Comm-F1 $\uparrow$} \\
\midrule
baseline\_all & 1  & 45.09  & 0.107 & 0.329 & 0.212 & 0.043 & 0.090 & 0.787 & 0.239 & — \\
zero\_shot    & 5  & 57.99 & 0.263 & 0.734 & 0.192 & 0.065 & 0.302 & 0.829 & 0.025 & — \\
S1 subreddit  & 10 & 38.29  & 0.083 & 0.296 & 0.216 & 0.104 & 0.067 & 0.792 & 0.079 & 0.262 \\
S2 graph      & 10 & 40.60  & 0.097 & 0.318 & 0.217 & 0.055 & 0.078 & 0.790 & 0.105 & 0.239 \\
S3 semantic   & 10 & 48.95  & 0.117 & 0.346 & 0.189 & 0.055 & 0.073 & 0.803 & 0.205 & 0.132 \\
S4 hybrid     & 10 & 42.58  & 0.106 & 0.334 & 0.228 & 0.058 & 0.113 & 0.799 & 0.157 & 0.224 \\
S5 interaction & 10 & 43.52 & 0.113 & 0.346 & 0.216 & 0.048 & 0.113 & 0.795 & 0.164 & 0.167 \\
\bottomrule
\end{tabular} 
}
\end{table}

Table~\ref{tab:strategy_comparison} reports per-adapter generation quality on \textsc{IBM Granite 4.1-3B}. All five community-conditioned adapters reduce perplexity by 16--34\% relative to the zero-shot base model and lower sentiment divergence by roughly $3$--$4\times$ (Sent-JSD 0.07--0.11 vs.\ 0.30), confirming that LoRA fine-tuning on community-specific data captures lexical fluency and affective style. The strongest persona signal comes from S1 (subreddit), which attains the lowest perplexity (38.3), the lowest sentiment-JSD (0.067), and the highest community classifiability (Comm-F1 $=0.262$, $2.6\times$ chance over 10 classes). Crucially, the Comm-F1 ranking across strategies (S2 $>$ S4 $>$ S5 $>$ S3) mirrors the NMI-vs-baseline ranking in Table~\ref{tab:grouping}, providing downstream confirmation that agreement with subreddit identity predicts behavioral identifiability. We further observe a trade-off between \emph{identifiability} and \emph{distributional similarity to real text}: S3 (semantic) achieves the best MAUVE and BERTScore-F1 but the weakest Comm-F1, whereas S1 is the inverse. Finally, because S4 is an edge-reweighted addition of S2, the two strategies differ only modestly on most metrics; the near-identity of their downstream metrics motivates richer hybridization schemes as future work.

\section{Conclusions and Future Work}
\label{sec:conclusion}

We presented \textsc{RedditPersona}, a modular framework that turns Reddit posts and comments into community-conditioned language model adapters, along with metrics for comparing them. Across 112 subreddits in the urban well-being domain, we compared five grouping strategies and found that the Comm-F1 ranking of adapters mirrors the NMI ranking of grouping strategies versus the subreddit baseline, confirming that partition quality predicts behavioral identifiability. We also observed a consistent trade-off between identifiability and distributional similarity to real text: subreddit-conditioned adapters are the most identifiable but least natural, while semantic adapters are the most natural but least identifiable.

Several directions remain open. Immediate research applications include studying discourse norms across domains beyond urban well-being, generating community-conditioned training data for social science simulations, and benchmarking community grouping assumptions in polarization or public health settings. Extending the framework to other model families would test whether the metrics hold across architectures. Richer hybridization schemes beyond linear graph-semantic blending, as well as the design of new grouping strategies, are major directions for future work. Finally, contrasting community-level adapters with individual-level fine-tuning would directly quantify the practical value of community abstraction for persona simulation research.

\begin{acks}
This work was supported by the Emerging Projects program, Infotech Oulu; the Research Council of Finland through the 6G Flagship program (grant 318927); the Strategic Research Council affiliated with Academy of Finland through the CO2CREATION project (grant 372355); European Commission (101137711); Business Finland through the Neural pub/sub research project (diary number 8754/31/2022); and the ERDF (project numbers A81568, A91867).
\end{acks}

\section*{Generative AI Usage Disclosure}
We used Generative AI (OpenAI GPT-5.5 and Claude Opus 4.7) strictly for proofreading: grammar correction, minor word-choice substitutions, and limited rephrasing of sentences we had already drafted. The tool did \emph{not} generate new scientific content, claims, or paragraphs. We used GitHub Copilot in Visual Studio Code to surface completion suggestions while cleaning and reorganizing our Python utilities. All suggestions were manually reviewed; no AI-generated code fragments remain in the final artifact.

\bibliographystyle{ACM-Reference-Format}
\bibliography{references}

\end{document}